\documentclass{article}

  \usepackage{iclr2027_conference}

  \usepackage{times}
  \usepackage{url}
  \usepackage{amsmath}
  \usepackage{amssymb}
  \usepackage{graphicx}
  \usepackage{booktabs}
  \usepackage{microtype}
  \usepackage{xcolor}
  \usepackage{hyperref}
  \usepackage{enumitem}
  \usepackage{multirow}       % multirow
  \usepackage{makecell}
  \usepackage{graphicx}
  \usepackage[table]{xcolor}
  \usepackage{bm}
  \title{1Step-WAM: Your World Action Model but Faster using 1-Step Action Generation}

\iclrfinalcopy

\author{%
  \textbf{Rundong Su$^{1*}$ \quad
  Pengtao Chen$^{1*}$ \quad
  Haohan Geng$^{2*}$ \quad
  Sha Zhang$^{2}$} \\
  \textbf{Zhihang Yuan$^{3}$ \quad
  Yicheng Rong$^{1}$ \quad
  Chong Yu$^{1}$ \quad
  Hao Huang$^{1}$ \quad
  Tao Chen$^{1,4\dagger}$} \\[4pt]
  \normalfont
  $^{1}$Fudan University \quad
  $^{2}$University of Science and Technology of China  \\
  $^{3}$Peking University \quad
  $^{4}$Shanghai Innovation Institute
}

\begin{document}

  \maketitle
  
    \begingroup
        \renewcommand{\thefootnote}{\fnsymbol{footnote}}
        \footnotetext[1]{Equal contribution.\hspace{1em}
        \textsuperscript{$\dagger$}Corresponding author.}
    \endgroup
  
  \begin{abstract}
  World Action Models (WAMs) unify visual world modeling and action generation, yet their real-world deployment is heavily bottlenecked by iterative action denoising and the sequential execution of visual prefilling and action prediction. 
We demonstrate that the action-denoising trajectories in WAMs are approximately straight, rendering the first-step velocity a strong estimate of the final action direction. 
Building on this observation, we introduce 1Step-WAM, a framework that transforms multi-step action denoising into one single forward. 
Specifically, we employ a lightweight calibrator to directly correct the first step velocity, and use bandwidth-constrained action decoding to further reduce errors in the calibrated one-step actions.
Compressing action generation into a single step enables it to execute in parallel with visual prefilling, completely breaking the sequential bottleneck of the traditional imagination-then-action paradigm.
Evaluations on RoboTwin and LIBERO show that 1Step-WAM largely preserves multi-step task success with only about 13.4M trainable parameters, while achieving inference speedups of $7.44\times$, $5.98\times$, and $24.84\times$ for FastWAM, ImageWAM and Kairos, respectively.
Furthermore, real-world robotic deployments using the FastWAM backbone confirm a consistent 7.44$\times$ speedup while maintaining close success rate to the multi-step baseline. Code is avaliable at  \url{https://github.com/changdong1687/1Step-WAM}.

  \end{abstract}

  \section{Introduction}
\label{sec:introduction}
% 第一段：WAM背景，说明他很强，应用广等等。第二段：WAM目前的问题（主要是效率上）点名两个问题。第三段：针对目前的问题提出解决方案。第四段：得到的实验结果简述。

In the field of Embodied AI, making robot policy capable of generating right actions across diverse environments remains a long-standing  objective. 
In the past two years, Vision-Language-Action (VLA) models \citep{kim2024openvla, brohan2023rt2, black2024pi0, intelligence2025pi05, bjorck2025gr00t, shukor2025smolvla, team2025gemini, team2025galaxea}have become a widely used model architecture in Embodied AI, it utilizes a pretrained vision-language-model (VLM) to generate  the actions from current visual observations and language instructions. 
However, VLA models adopt static VLM visual understanding ability, it lacks the ability of  understanding how  physical world will change after interaction, limiting their generalization ability. 
This has motivated the emergence of WAMs \citep{yuan2026fastwam, zhang2026imagewam, li2026causal}, which unify future prediction and action generation in a single model, exhibiting better generalization ability than VLAs.

Despite WAMs' superior generalization, the real-world deployment of WAMs is significantly bottlenecked by inference latency. 
The standard WAM pipeline relies on a sequential imagination-then-action process, they first imagine a future scence using a visual diffusion models~\citep{wan2025, hong2022cogvideo, labs2025flux, ali2025world} through multi step denoise, then denoise the actions based on the imagined future.
While recent optimization efforts have successfully accelerated the imagination phase by reducing the video denoising process to a single-step visual prefill, they largely overlook the latency bottleneck within the action generation phase itself. 
In these existing frameworks, a visual Diffusion Transformer (DiT) ~\citep{peebles2023scalable} first processes observations and instructions to extract visual KV caches, after which an action DiT must still undergo a time-consuming, multi-step iterative denoising process to produce control signals. 
This leaves two critical efficiency issues unresolved: the inherent latency of iterative action denoising, and the serial dependency that forces action generation to stall until the visual prefill is fully completed. 
This sequential mechanism fundamentally blocks the potential for parallel execution, thereby restricting the inference speed necessary for high-frequency robotic control~\citep{black2026real}.

To address these challenges, we propose 1Step-WAM, a novel framework that realizes one-step, parallel action generation for WAMs. 
% Simply replacing a multi-step solver with one Euler update is unreliable because a velocity field trained for local integration need not support a direct jump from noise to action. 
% However, 
We find that when the visual observation, language instruction, and robot state are fixed, the target action is nearly invariant to the initial noise used for generation. 
Building on this observation, we theoretically characterize the denoising dynamics and show that the action trajectory is approximately straight: the first-step velocity provides a strong estimate of the final action direction. 
Exploiting this property, we design a lightweight calibrator attached to the action DiT. This calibrator learns to directly correct the initial predicted velocity toward the multi-step teacher trajectory, effectively transforming the computationally expensive iterative solver into a direct noise-to-action one-step policy.
Furthermore, to mitigate the remaining action deviation between our one-step policy and the multi-step teacher, we analyze the generated actions in the frequency domain. 
 Per-joint DCT visualizations in Appendix~\ref{app:action_spectra} show that teacher action concentrates most of their spectral energy in low-frequency modes. This observation motivates bandwidth-constrained action decoding: we apply a temporal DCT along the action horizon and retain only the first $K$ modes. 
 The projection trades the loss of high-frequency teacher components against the removal of prediction error in the discarded frequencies. 
 Our analysis shows that this projection reduces the expected squared error relative to the teacher when the discarded prediction-error energy exceeds the discarded teacher-action energy.
Most importantly, compressing the action generation into a single step unlocks a critical system-level breakthrough. 
Instead of waiting for the visual prefill to finish, 1Step-WAM integrates action generation directly into the visual prefilling computation, eliminating the sequential waiting time of the traditional imagination-then-action paradigm.

\begin{figure}[t]
    \centering
    \includegraphics[width=\linewidth]{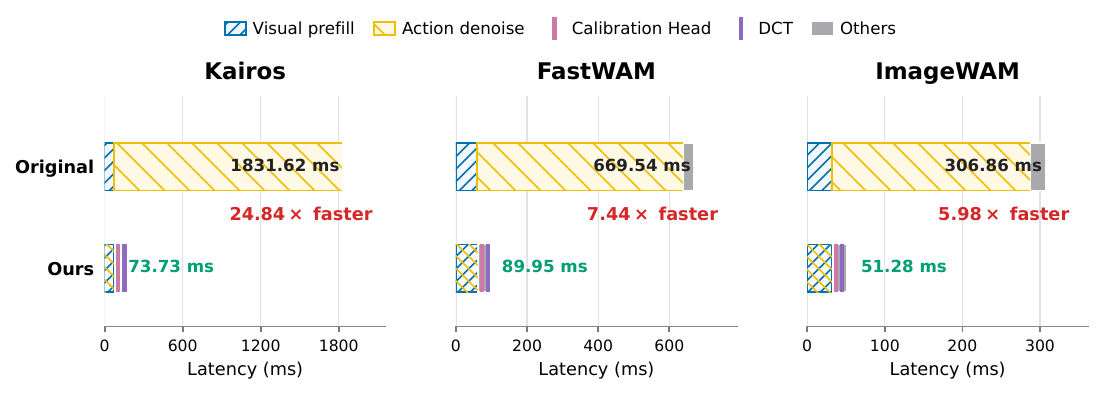}
    \caption{Speedups of 1Step-WAM over the original multi-step FastWAM, ImageWAM, and Kairos.}
    \label{fig:intro_latency_breakdown}
\end{figure}

We comprehensively evaluate our proposed 1Step-WAM in both simulated and real-world tasks. 
To demonstrate its broad applicability, we apply our method to three WAM models: FastWAM, ImageWAM, and Kairos. 
Our approach achieves remarkable speedups of $7.44\times$ for FastWAM, $5.98\times$ for ImageWAM, and $24.84\times$ for Kairos, as shown in Fig.~\ref{fig:intro_latency_breakdown}, all while suffering a negligible drop in task success rate on RobotWin~\citep{chen2025robotwin} and Libero~\citep{liu2023libero} benchmark.
Furthermore, we validate 1Step-WAM on real robot tasks using the FastWAM backbone. In these real-world deployments, our method consistently delivers a $7.44\times$ acceleration and maintains an average task success rate 75\%, compared to the 80\% of the original multi-step denoise baseline.
% These empirical results compellingly demonstrate that OneStep-WAM successfully bridges the gap between computationally heavy predictive world models and the stringent real-time requirements of practical robotic control.

Our main contributions can be summarized as follows:
\begin{itemize}
    \item We demonstrate that WAM action trajectories are approximately straight and highly invariant to initial noise. This establishes that the first-step predicted velocity inherently includes the primary direction of the final target action.
    \item We propose 1Step-WAM, a framework combines lightweight velocity calibration, bandwidth-constrained action decoding, and joint video-action execution for faster action generation.
    \item We evaluate 1Step-WAM across FastWAM, ImageWAM, and Kairos on RoboTwin and LIBERO, and on real-robot tasks with FastWAM,  it achieves end-to-end inference speedups of $7.44\times$, $5.98\times$, and $24.84\times$ respectively, while largely preserving task success.
\end{itemize}

  \section{Related Work}
\label{sec:related_work}

\subsection{World Action Models}
\label{subsec:world_action_models}

World Action Models (WAMs) use visual world modeling to support action generation.
A central design choice is how predicted visual representations are used during inference.
Recent approaches reduce the dependence on explicit future generation at deployment.
Fast-WAM~\citep{yuan2026fastwam} and GigaWorld-Policy~\citep{ye2026gigaworld} retain visual-dynamics supervision during training while enabling inference without future-video synthesis. ImageWAM instead conditions action generation on features from an image-editing model~\citep{zhang2026imagewam}, while Kairos~\citep{kairosteam2026kairosregretawarenativeworldaction} integrates
understanding, generation, and prediction within a unified framework.

These approaches reduce the cost of visual world modeling, but action generation still requires iterative denoising. We address this bottleneck by converting pretrained multi-step
  action generators into one-step policies. This approach complements visual world modeling acceleration.

\subsection{Diffusion Distillation}
\label{subsec:diffusion_distillation}

Diffusion distillation reduces the sequential network evaluations for sampling. Representative approaches
include progressive step reduction~\citep{salimans2022progressive}, trajectory consistency~\citep{song2023consistency}, distribution matching~\citep{yin2024onestep,yin2024improved}, and step-size-conditioned shortcuts~\citep{frans2025shortcut}.
% These ideas also enable faster robot control, 
In robotics, Consistency Policy~\citep{prasad2024consistency} and OneDP~\citep{wang2024onedp} apply consistency and distribution matching to diffusion policies.
One-Step Flow Policy uses self-distillation for flow-based control~\citep{li2026onestepflow}, while SnapFlow~\citep{luan2026snapflow} progressively distills flow-matching VLAs into one-step action generators. Most closely, Flash-WAM~\citep{akbari2026flash} studies step distillation directly in WAMs, it propose modality-specific consistency parameterizations and directly distills action DiT  into one step denoise.

However, all these methods need to train the whole dit params which is expensive for training. 
Our method focuses on action generation in pretrained WAMs: we freeze the backbone and train only a lightweight component to calibrate the initial velocity toward the multi-step teacher endpoint.

  \section{Rethinking: Can a WAM Generate Actions in a Single Step?}
\label{subsec:Rethinking}

In continuous action generation for WAMs, action DiT typically require multi-step iterative denoising to generate an action chunk. 
However, we observe a critical phenomenon: when visual observations, language instructions and robot state are fixed, the final action distribution decoded through multiple steps exhibits extremely small variance, regardless of the initial noise at the starting point of denoising.
We run FastWAM's multi-step denoiser with 10,000 independently sampled initial noises. The resulting actions have an empirical variance of $1.182\times10^{-5}$.
This phenomenon inspires us to rethink: \textit{Can a WAM generate usable actions in a single forward pass?}

To mathematically answer this question, we conduct a theoretical derivation from the trajectory geometry of Flow Matching~\citep{lipman2022flow}. 
For fixed conditioning inputs, we assume the clean target action \( \bm x_0 \sim \mathcal{N}(\bm \mu, \sigma^2 \bm I_d)\), with \(\sigma>0\), and the pure noise \(\bm x_1 \sim \mathcal{N}(\bm 0, \bm I_d)\). The forward noising path is defined as \( \bm x_t = (1-t)\bm x_0 + t \bm x_1\), where \(t\in[0,1]\); denoising proceeds in the reverse direction, from \(t=1\) to \(t=0\). Since the clean target \(\bm x_0\) and the noise \(\bm x_1\) are sampled independently for this interpolation, the marginal variance of the intermediate state \(\bm x_t\) is:
\begin{equation}
\text{Var}(\bm x_t) = (1-t)^2 \text{Var}(\bm x_0) + t^2 \text{Var}(\bm x_1) = \left[ (1-t)^2\sigma^2 + t^2 \right] \bm I_d
\end{equation}

For clarity, we denote this time-dependent variance scalar as \(D(t) = (1-t)^2\sigma^2 + t^2\). In the flow matching framework, the optimal vector field is defined with respect to increasing \(t\) as the conditional expectation of the interpolation velocity given the current state:
\begin{equation}
\bm v^*(\bm x_t, t) = \mathbb{E}[\bm x_1 - \bm x_0 \mid \bm x_t]
\end{equation}

Since the independently sampled endpoints are jointly Gaussian, this conditional expectation can be evaluated analytically. Using \(\mathbb{E}[\bm x_1-\bm x_0]=-\bm \mu\), the marginal mean \(\mathbb{E}[\bm x_t] = (1-t)\bm \mu\), the variance \(D(t)\bm I_d\), and the cross-covariance \(\text{Cov}(\bm x_1 - \bm x_0, \bm x_t) = [t - (1-t)\sigma^2]\bm I_d\), the Gaussian conditional expectation identity gives:
\begin{equation}
\begin{aligned}
\bm v^*(\bm x_t, t)
&= -\bm \mu + \frac{t - (1-t)\sigma^2}{D(t)}\bigl[\bm x_t-(1-t)\bm \mu\bigr] \\
&= -\frac{t}{D(t)}\bm \mu + \frac{t - (1-t)\sigma^2}{D(t)}\bm x_t
\end{aligned}
\end{equation}

During inference, the system integrates the ordinary differential equation (ODE) \(\dot{\bm x}_t = \bm v^*(\bm x_t, t)\) backward from \(t=1\) to \(t=0\), with initial noise \(\bm x(1)=\bm x_1\).
Here, $\bm v^*(\bm x,t)$ specifies the velocity at any state $\bm x$, whereas $\dot{\bm x}_t$ is the derivative of the particular trajectory with respect to increasing $t$; a denoising step uses $\Delta t<0$. Below, $\bm x_t$ denotes this inference trajectory, which shares the marginals of the training interpolation but couples its endpoints through the ODE. By analytically solving this first-order linear ODE, we obtain:
\begin{equation}
\bm x_t = \sqrt{D(t)} \bm x_1 + (1-t)\bm \mu = \sqrt{(1-t)^2\sigma^2 + t^2} \, \bm x_1 + (1-t)\bm \mu
\end{equation}
At clean endpoint, this gives \(\bm x_{t=0}=\bm \mu+\sigma \bm x_1\), which has the assumed target distribution \(\mathcal{N}(\bm \mu,\sigma^2 \bm I_d)\).

Taking the derivative of this analytical trajectory with respect to \(t\) gives:
\begin{equation}
\dot{\bm x}_t = -\bm \mu + C(t) \bm x_1, \quad \text{where } C(t) = \frac{t - (1-t)\sigma^2}{\sqrt{(1-t)^2\sigma^2 + t^2}}
\end{equation}

The trajectory velocity exhibits two regimes as $\sigma \to 0^+$:
\begin{itemize}
    \item \textbf{Collinear Main Interval:} For any fixed \(\epsilon\in(0,1)\), on the time interval \(t \in [\epsilon,1]\), the time-dependent noise coefficient \(C(t)\) uniformly converges to \(1\). Thus, \(\dot{\bm x}_t\) approaches the constant vector \(\bm x_1-\bm \mu\). Since denoising proceeds with decreasing \(t\), its direction approaches \(\bm \mu-\bm x_1\), and the trajectory approaches a straight line over this interval.
    \item \textbf{Terminal transition:} Within the narrow terminal interval $t=\mathcal{O}(\sigma)$ near the clean endpoint, the coefficient $C(t)$ departs from $1$, reaching $C(0)=-\sigma$. Thus, $\dot{\bm x}_{t=0}=-\bm \mu-\sigma \bm x_1\to-\bm \mu$ as $\sigma\to0^+$; the reverse-time denoising velocity $-\dot{\bm x}_t$ approaches $\bm \mu$ at this endpoint.
\end{itemize}

Thus, under the small-variance Gaussian model, a velocity prediction at the noise endpoint \(t=1\) identifies the primary denoising direction \(\bm \mu-\bm x_1\). With this time convention, a one-step Euler estimate is \(\widehat{\bm x}_0=\bm x_1-\bm v^*(\bm x_1,1)\), where the minus sign follows from the time step \(\Delta t=-1\).

  \section{Methods}
\begin{figure*}[t]
    \centering
    \includegraphics[width=\textwidth]{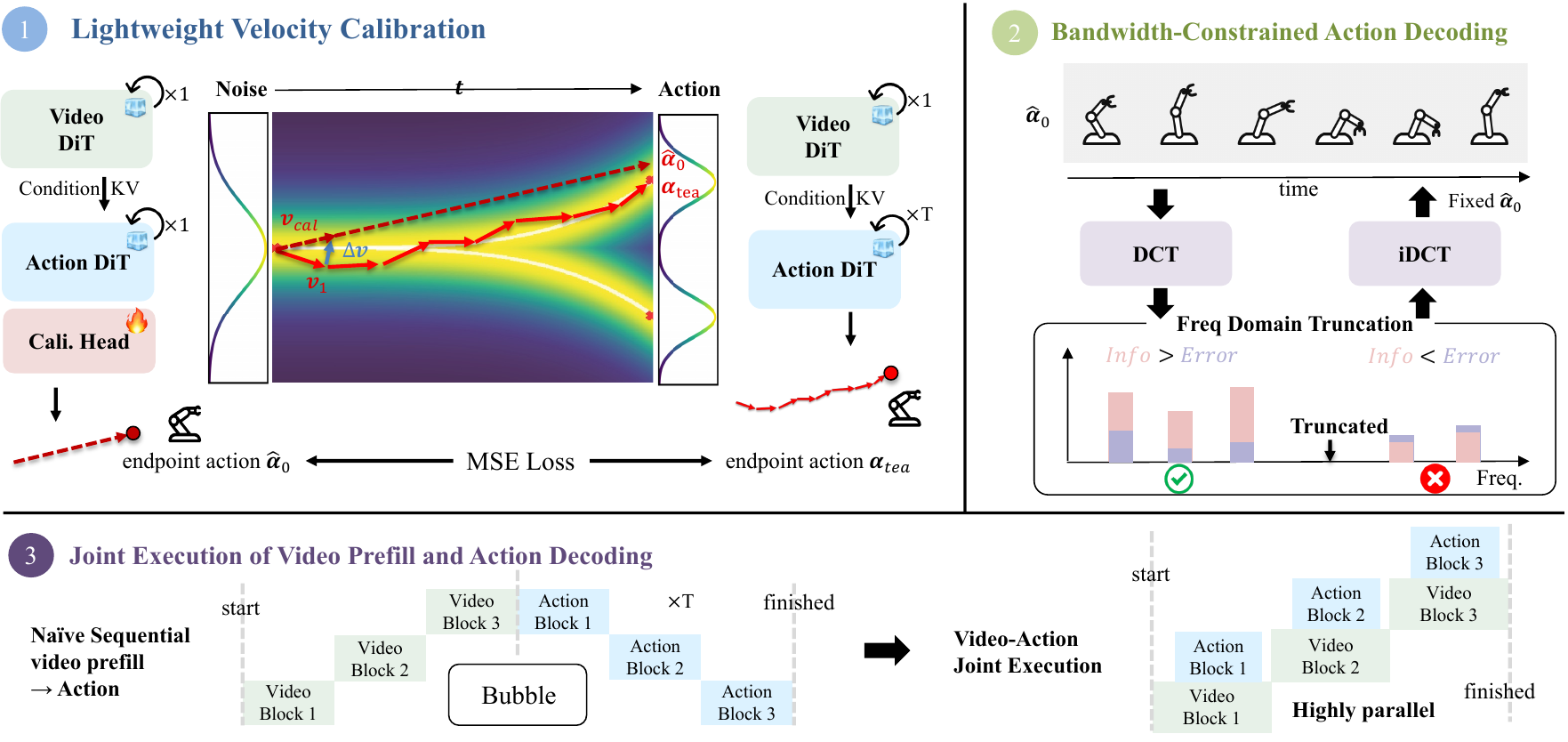}
    \caption{Overview of 1Step-WAM.  It combines lightweight velocity calibration, bandwidth-constrained action decoding, and joint execution of video prefill for faster action generation.}
    \label{fig:method_overview}
\end{figure*}
\subsection{Lightweight Velocity Calibration}

Motivated by the nearly linear denoising trajectory, we introduce a lightweight velocity calibration head to calibrate the initial velocity for one-step action prediction.

\textbf{Calibration Inputs and Formulation.} 
Let \(\bm x_1 \sim \mathcal{N}(\bm 0, \bm I_d)\) denote the initial noise. In our  framework, we execute the frozen action DiT backbone only once at \(t=1\). This single forward pass yields the raw velocity prediction \(\bm v_{1}\), alongside the intermediate hidden features \(\bm h_1\) last action dit block output. 

To compensate for the initial regression error inherent in predicting from pure noise, we introduce a MLP-based calibration head, denoted as \(f_\phi\). This head utilizes the priors informations to predict a velocity residual \(\Delta \bm v\):
\begin{equation}
    \Delta \bm v = f_\phi(\bm h_1, \bm x_1, \bm v_{1})
\end{equation}
The calibrated effective velocity can be formulated as \(\bm v_{cal} = \bm v_{1} + \Delta \bm v\). Consequently, the one-step action can be expressed as:
  \begin{equation}
      \hat{\bm a}_0 = \bm x_1 - \left(\bm v_1 + f_\phi(\bm h_1, \bm x_1, \bm v_1)\right).
  \end{equation}
where \(\hat{\bm a}_0\) represents the final generated action sequence.

\textbf{Calibration Objective.}
The training objective for the calibration head is to minimize the discrepancy between the one-step generated action \(\hat{\bm a}_0\) and the multi-step teacher action \(\bm a_{tea}\).
During training, we freeze the WAM backbone, and only update the calibration head parameters \(\phi\). The training objective is formulated as a Mean Squared Error (MSE) loss directly in the action space:
\begin{equation}
    \mathcal{L} = \mathbb{E} \left[ \left\| \hat{\bm a}_0 - \bm a_{tea} \right\|_2^2 \right]
\end{equation}
By minimizing \(\mathcal{L}\), the calibration head implicitly learns the non-linear transformation required to map the raw initial velocity to an effective velocity that better matches the teacher endpoint. 

\subsection{Bandwidth-Constrained Action Decoding}

Velocity calibration learns to approximate the multi-step teacher
endpoint with a single action DiT forward. The endpoint-matching
objective, however, does not explicitly constrain the temporal
bandwidth of the predicted action chunk. We complement calibration
with a fixed projection onto a low-frequency temporal subspace.
This projection trades the loss of teacher motion in discarded
frequencies against the removal of prediction error in those
frequencies.

\paragraph{Temporal projection.}
Let $\widehat{\bm A}_{c} \in \mathbb{R}^{T \times d_c}$ denote the
calibrated one step action, and let
$\bm C \in \mathbb{R}^{T \times T}$ be the orthonormal DCT-II matrix.
We define $\bm M_K$ as a diagonal mask retaining the first $K$ modes,
including the constant mode. The final executable continuous action is
\begin{equation}
    \bm A^{\mathrm{exec}}_{c}
    = \bm P_K \widehat{\bm A}_{c},
    \qquad
    \bm P_K = \bm C^\top \bm M_K \bm C.
\end{equation}
Since $\bm P_K$ is an orthogonal projector, this operation returns
the closest trajectory to $\widehat{\bm A}_{c}$ within the retained
temporal subspace. The transform is applied independently to each joint along the action horizon, while the gripper remains unchanged.

\paragraph{When does truncation help?}
Let $\bm A^{\mathrm{T}}_{c}$ denote the multi-step teacher action
under the same conditioning and initial noise, and define mismatch error
$\bm E_c = \widehat{\bm A}_{c} - \bm A^{\mathrm{T}}_{c}$.
To quantify this trade-off, let $\mathcal{R}_K$ denote the overall expected residual error between the projected prediction and the teacher's action. 
By the orthogonality of the projection, $\mathcal{R}_K$ explicitly decomposes into two components:
\begin{equation}
\begin{aligned}
    \mathcal{R}_K
    &:= \mathbb{E}
    \left\|\bm P_K\widehat{\bm A}_{c}-\bm A^{\mathrm{T}}_{c}\right\|_F^2 \\
    &= \underbrace{\mathbb{E}\|\bm P_K \bm E_c\|_F^2}
       _{\text{retained-band prediction error}}
     + \underbrace{
       \mathbb{E}\|(\bm I-\bm P_K)\bm A^{\mathrm{T}}_{c}\|_F^2}
       _{\text{discarded teacher energy}}.
\end{aligned}
\end{equation}
Here, $\|\cdot\|_F$ denotes the Frobenius norm, with
$\|\bm X\|_F^2 = \sum_{t=1}^{T}\sum_{j=1}^{d_c} \bm X_{t,j}^2$,
and the expectation is taken over conditioning inputs and
initial noise.

Relative to the unprojected action, whose error is
$\mathcal{R}_{\mathrm{full}}=\mathbb{E}\|E_c\|_F^2$, we have
\begin{equation}
    \mathcal{R}_K-\mathcal{R}_{\mathrm{full}}
    =
    \mathbb{E}\|(\bm I-\bm P_K)\bm A^{\mathrm{T}}_{c}\|_F^2
    -
    \mathbb{E}\|(\bm I-\bm P_K)\bm E_c\|_F^2.
\end{equation}
Truncation therefore improves teacher approximation (i.e., $\mathcal{R}_K < \mathcal{R}_{\mathrm{full}}$) when the
discarded frequencies contain more prediction error than teacher
motion. Low-frequency concentration of teacher actions alone
does not guarantee this condition.

We train the calibrator with the full-band teacher-matching
objective described above and apply the fixed projection at
inference. The cutoff $K$ controls the trade-off between preserving
teacher motion and further suppressing prediction error, $K=T$ recovers
the unprojected prediction. 

\subsection{Joint Execution of Video Prefill and Action Denoising}

One-step calibration reduces action generation to a single action DiT forward pass. Under the conventional sequential schedule, however, this pass still waits for the entire video prefill to finish. 
During inference, the video branch produces visual key/value (K/V) features at each layer, which are cached and used to condition the corresponding action block. 
We exploit this layer-wise conditioning to start action computation as soon as the required visual features become available.

\paragraph{Layer-level dependencies.}
Starting action computation before video prefill finishes requires identifying the inputs needed by each action block. 
Let $\mathcal{C}_l$ denote the visual K/V cache produced by video
layer $l$, and let $\bm H^a_{l-1}$ denote the input hidden states
of action block $l$. The action block computes
\begin{equation}
    \bm H^a_l
    = \mathcal{A}_l\left(\bm H^a_{l-1}, \mathcal{C}_l\right),
\end{equation}
where $\mathcal{A}_l$ denotes the corresponding action DiT
block. Its execution requires both $H^a_{l-1}$ and
$\mathcal{C}_l$, but does not depend on the caches produced
by later video layers. Thus, completing the entire video
prefill is not a prerequisite for starting action computation.

\paragraph{Joint scheduling.}
The layer-level dependencies above permit a pipelined execution of video prefill and action generation.
We assign the video and action branches to separate CUDA streams and use CUDA events to synchronize access to the visual K/V caches.
After video layer $l$ produces $\mathcal{C}_l$, it records an event indicating that the cache is ready.
The action stream waits for this event before executing action block $l$, while the video stream proceeds to the next layer. 
Execution within the action stream preserves the ordering of action blocks. After the final action block, the calibration head and temporal projection produce the executable action chunk.
This schedule preserves the original computational dependencies while allowing video and action computation to overlap.

  \section{Experiments}
\label{sec:experiments}

\providecommand{\expTBD}{\textcolor{blue}{\texttt{TBD}}}
\providecommand{\expdetail}[1]{\textcolor{blue}{[TBD: #1]}}

In this section, we evaluate whether 1Step-WAM preserves task success while reducing inference latency. All latency is measured on one NVIDIA L40 GPU.
% Our experiments include two simulation benchmarks, a real-robot evaluation, an analysis of the conditional concentration underlying one-step prediction, and ablations of calibration inputs and temporal decoding.

\subsection{Experimental Setup}
\label{sec:experimental_setup}

\noindent \textbf{Benchmarks.}
We evaluate on two simulation benchmarks and a real-robot setup.
RoboTwin~\citep{chen2025robotwin} is a bimanual manipulation benchmark with clean and randomized for 50 tasks.
LIBERO~\citep{liu2023libero} provides complementary evaluations through four task suites: LIBERO-Spatial, LIBERO-Object,LIBERO-Goal, and LIBERO-Long, each containing 10 tasks. 
For these two task we following the evalation settings in FastWAM~\citep{yuan2026fastwam}, ImageWAM~\citep{zhang2026imagewam} and Kairos~\citep{kairosteam2026kairosregretawarenativeworldaction}
For real-world evaluation, we deploy FastWAM-based policies on an AIRBOT Play robotic arm, using a fisheye camera for visual observations. We evaluate three manipulation tasks:
(1) \textit{Pick Place}: pick up all the fruits and vegetables on the table and place them in the basket, (2) \textit{Stack Cups}: stack all the cups on the table, and
(3) \textit{Hang Mug}: hang all the mugs on the table. The training set contains 200 episodes per task, and we train the policies for five epochs. 
During evaluation each task is conducted 20 independent rollouts per task per method and report task-level and average success rates.

\noindent \textbf{Models and Baselines.}
We apply 1Step-WAM to FastWAM~\citep{yuan2026fastwam}, ImageWAM~\citep{zhang2026imagewam}, and Kairos~\citep{kairosteam2026kairosregretawarenativeworldaction}.
On RoboTwin, we additionally compare with DMD2~\citep{yin2024improved} distillation model used in video DiTs, SnapFlow~\citep{luan2026snapflow}, an action-distillation method developed for VLAs and Flash-WAM~\citep{akbari2026flash} distillation method developed for WAMs.
We adapt these methods to each backbone action branch. For fair comparison, all methods use the same observation interface, action representation and horizon within each backbone.

\noindent \textbf{Implementation Details.}
We freeze the pretrained WAM backbone and train only the velocity calibration head for 500 steps using the AdamW optimizer, with a learning rate of $5\times10^{-6}$ and a global batch size of 512. The $K$ used for FastWAM~\citep{yuan2026fastwam}, ImageWAM~\citep{zhang2026imagewam} and Kairos~\cite{kairosteam2026kairosregretawarenativeworldaction} is 8,4,8.

\subsection{Main Results}
\label{sec:simulation_results}

\paragraph{Simulation Benchmarks.}
Table~\ref{tab:robotwin_main} reports success rates and inference latency on RoboTwin~\citep{chen2025robotwin}. Across all three backbones, 1Step-WAM achieves the highest success rate among the evaluated one-step methods in both settings.
Its average success rates are 90.89\%, 92.44\%, and 95.20\% on FastWAM, ImageWAM, and Kairos, respectively, only 0.94, 0.90, and 0.76 percentage points below their multi-step counterparts. 
At the same time, end-to-end inference is accelerated by $7.44\times$, $5.98\times$, and $24.84\times$, respectively.
Despite outperforming both DMD2~\citep{yin2024improved}, SnapFlow~\citep{luan2026snapflow} and Flash-WAM~\citep{akbari2026flash} in RoboTwin success rate across all three backbones, 
1Step-WAM trains only 13.4M parameters, using much fewer trainable parameters than either distillation baseline.

Table~\ref{tab:libero_main} reports success rates across the four LIBERO suites. 1Step-WAM closely matches its multi-step counterpart on FastWAM and ImageWAM, with mean success rates of 96.6\% versus 97.4\% and 98.2\% versus 98.4\%, respectively. On Kairos, the one-step model reaches 97.2\%, exceeding the multi-step baseline by 1.0 percentage point. Compared with naive one-step generation, calibration improves mean success by 0.9, 1.4, and 2.0 percentage points on the three backbones, respectively.

% Table~\ref{tab:libero_main} compares each multi-step backbone with its calibrated one-step counterpart on LIBERO, demonstrating the robutness of our proposed methods.

\begin{table}[t]
    \centering
    \caption{
        Evaluation result on RoboTwin~\citep{chen2025robotwin} simulation benchmark
    }
    \label{tab:robotwin_main}
    \small
    \resizebox{\linewidth}{!}{%
    \begin{tabular}{@{}llcccccc@{}}
        \toprule[1.5pt]
        % \textbf{Backbone} & \textbf{Method} & \shortstack{\textbf{Train}\\\textbf{params}.}
        % & \textbf{Clean} & \textbf{Random} & \textbf{Avg.}
        % & \shortstack{\textbf{Latency}\\\textbf{(ms)}$\downarrow$}
        % & \textbf{Speedup} $\uparrow$ \\
            \multirow{2}{*}{\textbf{Model}}
        & \multirow{2}{*}{\textbf{Method}}
        & \multirow{2}{*}{\makecell{\textbf{Train}\\\textbf{params.}}}
        & \multirow{2}{*}{\textbf{Clean}}
        & \multirow{2}{*}{\textbf{Random}}
        & \multirow{2}{*}{\textbf{Avg.}}
        & \multirow{2}{*}{\makecell{\textbf{Latency}\\\textbf{(ms)}$\downarrow$}}
        & \multirow{2}{*}{\makecell{\textbf{Speedup}\\$\uparrow$}} \\
        & & & & & & & \\
        \midrule

        \multirow{5}{*}{\makecell{\textbf{FastWAM}\\\citep{yuan2026fastwam}}}
        % \multirow{5}{*}{\textbf{FastWAM}~\citep{yuan2026fastwam}}
        & Original
        & --
        & 91.88 & 91.78 & 91.83
        & 669.54 & $1.00\times$ \\

        & Naive 1-Step
        & --
        & 69.92 & 67.52 & 68.72
        & 147.62 & $4.54\times$ \\

        & DMD2~\citep{yin2024improved}
        & 2.04B
        & 82.48 & 83.40 & 82.94
        & 147.62 & $4.54\times$ \\

        & SnapFlow~\cite{luan2026snapflow}
        & 1.02B
        & 90.52 & 86.96 & 88.74
        & 147.69 & $4.53\times$ \\

        & Flash-WAM~\cite{akbari2026flash}
        & 1.02B
        & 90.44 & 89.80 & 90.12
        & 147.62 & $4.54\times$ \\

        \rowcolor[HTML]{EFEFEF}
        \cellcolor{white}& Ours (1Step-WAM)
        & \textbf{13.42M}
        & \textbf{91.10} & \textbf{90.68} & \textbf{90.89}
        & \textbf{89.95} & \textbf{7.44}$\times$ \\

        \midrule
        \multirow{5}{*}{\makecell{\textbf{ImageWAM}\\\citep{zhang2026imagewam}}}
        & Original
        & --
        & 93.38 & 93.30 & 93.34
        & 306.85 & $1.00\times$ \\

        & Naive 1-Step
        & --
        & 37.20 & 34.53 & 35.87
        & 77.64 & $3.95\times$ \\

        & DMD2~\citep{yin2024improved}
        & 1.28B
        & 64.52 & 65.40 & 64.96
        & 77.64 & $3.95\times$ \\

        & SnapFlow~\cite{luan2026snapflow}
        & 643.34M
        & 88.00 & 88.24 & 88.12
        & 77.69 &  $3.95\times$ \\

        & Flash-WAM~\cite{akbari2026flash}
        & 642.03M
        & 91.56 & 91.04 & 91.30
        & 77.64 & $3.95\times$ \\

        \rowcolor[HTML]{EFEFEF}
        \cellcolor{white}& Ours (1Step-WAM)
        & \textbf{13.40M}
        & \textbf{92.26} & \textbf{92.62} & \textbf{92.44}
        & \textbf{51.28} & \textbf{5.98}$\times$ \\

        \midrule

       \multirow{5}{*}{\makecell{\textbf{Kairos}\\\citep{kairosteam2026kairosregretawarenativeworldaction}}}
        & Original
        & --
        & 96.96 & 94.96 & 95.96
        & 1831.62 & $1.000\times$ \\

        & Naive 1-Step
        & --
        & 55.96 & 61.00 & 58.48
        & 133.81 & $13.69\times$ \\

        & DMD2~\citep{yin2024improved}
        & 1.90B 
        & 70.12 & 72.00 & 71.06
        & 133.81 & $13.69\times$ \\

        & SnapFlow~\cite{luan2026snapflow}
        & 0.96B
        & 85.08 & 82.28 & 83.68
        & 133.88 & $13.68\times$ \\

        & Flash-WAM~\cite{akbari2026flash}
        & 0.95B
        & 95.84 & 92.76 & 94.30
        & 133.81 & $13.69\times$ \\
        
        \rowcolor[HTML]{EFEFEF}
        \cellcolor{white}& Ours (1Step-WAM)
        & \textbf{13.42M}
        & \textbf{96.56} & \textbf{93.84} & \textbf{95.20}
        & \textbf{73.73} & \textbf{24.84}$\times$ \\

        \bottomrule[1.5pt]
    \end{tabular}}
\end{table}

\begin{table}[t]
    \centering
    \caption{
        Success rates (\%) on the LIBERO suites.
    }
    \label{tab:libero_main}
    \small
    \setlength{\tabcolsep}{5pt}
    
    \renewcommand{\arraystretch}{1.1}
    \begin{tabular}{@{}llrrrrr@{}}
        \toprule
        \textbf{Model} & \textbf{Method}
        & \textbf{Spatial} & \textbf{Object} & \textbf{Goal} & \textbf{Long} & \textbf{Mean} \\
        \midrule
        \multirow{3}{*}{\makecell{\textbf{FastWAM}\\\citep{yuan2026fastwam}}}
        & Original
        & 97.8 & 99.8 & 96.0 & 96.0 & 97.4 \\
        & Naive 1-Step
        & 97.0 & \textbf{99.2} & 95.0 & 91.6 & 95.7 \\
        &  \cellcolor[HTML]{EFEFEF}Ours (1Step-WAM)
        & \cellcolor[HTML]{EFEFEF}\textbf{97.4} & \cellcolor[HTML]{EFEFEF}\textbf{99.2} &\cellcolor[HTML]{EFEFEF} \textbf{95.4}
        & \cellcolor[HTML]{EFEFEF}\textbf{94.2} & \cellcolor[HTML]{EFEFEF}\textbf{96.6} \\
        \midrule
        \multirow{3}{*}{\makecell{\textbf{ImageWAM}\\\citep{zhang2026imagewam}}}
        & Original
        & 97.2 & 99.2 & 98.8 & 98.4 & 98.4 \\
        & Naive 1-Step
        & 95.8 & 99.6 & 97.2 & 94.6 & 96.8 \\
        & \cellcolor[HTML]{EFEFEF} Ours (1Step-WAM)
        & \cellcolor[HTML]{EFEFEF}\textbf{97.6} & \cellcolor[HTML]{EFEFEF}\textbf{100.0} & \cellcolor[HTML]{EFEFEF}\textbf{98.2}
        & \cellcolor[HTML]{EFEFEF}\textbf{96.8} & \cellcolor[HTML]{EFEFEF}\textbf{98.2} \\
        \midrule
        \multirow{3}{*}{\makecell{\textbf{Kairos}\\\citep{kairosteam2026kairosregretawarenativeworldaction}}}
        & Original
        & 97.4 & 99.2 & 97.0 & 91.2 & 96.2 \\
        & Naive 1-Step
        & 97.6 & \textbf{100.0} & 94.2 & 89.0 & 95.2 \\
        & \cellcolor[HTML]{EFEFEF} Ours (1Step-WAM)
        & \cellcolor[HTML]{EFEFEF}\textbf{99.0} & \cellcolor[HTML]{EFEFEF}99.0 & \cellcolor[HTML]{EFEFEF}\textbf{98.6}
        &\cellcolor[HTML]{EFEFEF} \textbf{92.2} & \cellcolor[HTML]{EFEFEF}\textbf{97.2} \\
        \bottomrule
    \end{tabular}
\end{table}

\begin{figure*}[t]
    \centering
    \includegraphics[width=\textwidth]{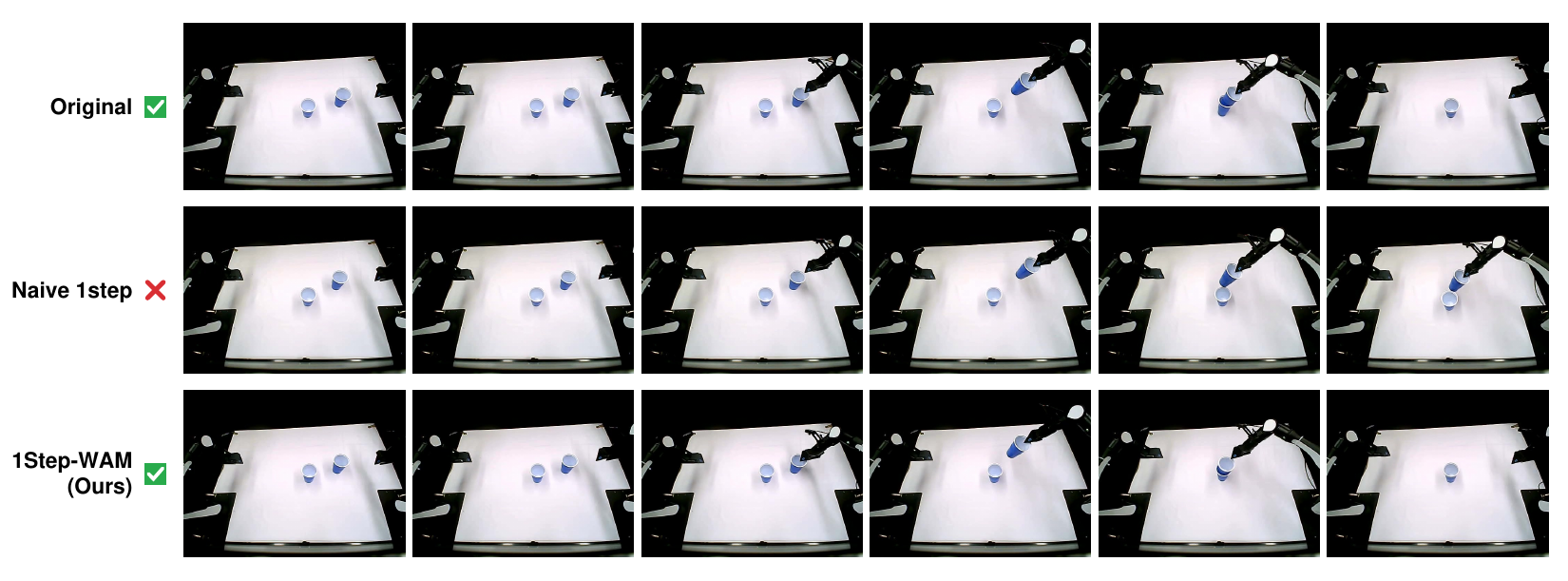}
    \caption{Real-world execution comparison on Stack Cups. In the illustrated rollouts, original FastWAM and 1Step-WAM successfully stack the cups, while naive 1-step fails.}
    \label{fig:real_world_tasks}
\end{figure*}

\paragraph{Real-World Task Evaluation.}
\label{sec:real_robot}
We evaluate FastWAM, naive one-step generation, and 1Step-WAM on an AIRBOT Play manipulator equipped with a fisheye camera. 
Table~\ref{tab:real_robot} reports their success rates. Pick Place tolerates relatively coarse motions, allowing even naive one-step generation to complete the task.
Stack Cups and Hang Mug, in contrast, require more precise alignment, making them more sensitive to action prediction errors.
Figure~\ref{fig:real_world_tasks} compares representative execution sequences of the three methods on Stack Cups. In the illustrated rollouts, both original FastWAM and 1Step-WAM successfully stack the cups, whereas naive one-step generation fails.
Additional execution comparisons on Pick Place and Hang Mug are provided in the appendix.
% Across the tasks, 1Step-WAM achieves 75\% average success, compared with 80\% for multi-step FastWAM and 40\% for naive one-step generation.

\begin{table}[t]
    \centering
    \caption{
        Real World Tasks success rates (\%) 
    }
    \label{tab:real_robot}
    \small
    \setlength{\tabcolsep}{6pt}
    \renewcommand{\arraystretch}{1.2}
    
    \begin{tabular}{@{}lrrrr@{}}
        \toprule
        \textbf{Method} & \textbf{Pick Place} & \textbf{Stack Cups} & \textbf{Hang Mug} & \textbf{Avg.} \\
        \midrule
        \textbf{Original FastWAM}
        & 100 & 90 & 50 & 80 \\
        \textbf{Naive 1-Step}
        & 100 & 40 & 10 & 50 \\
        \textbf{1Step-WAM}
        & \textbf{100} & \textbf{80} & \textbf{45} & \textbf{75} \\
        \bottomrule
    \end{tabular}
\end{table}

\subsection{Empirical Analysis of Denoising Trajectories}
\begin{figure*}[htbp]
    \centering
    \includegraphics[width=\textwidth]{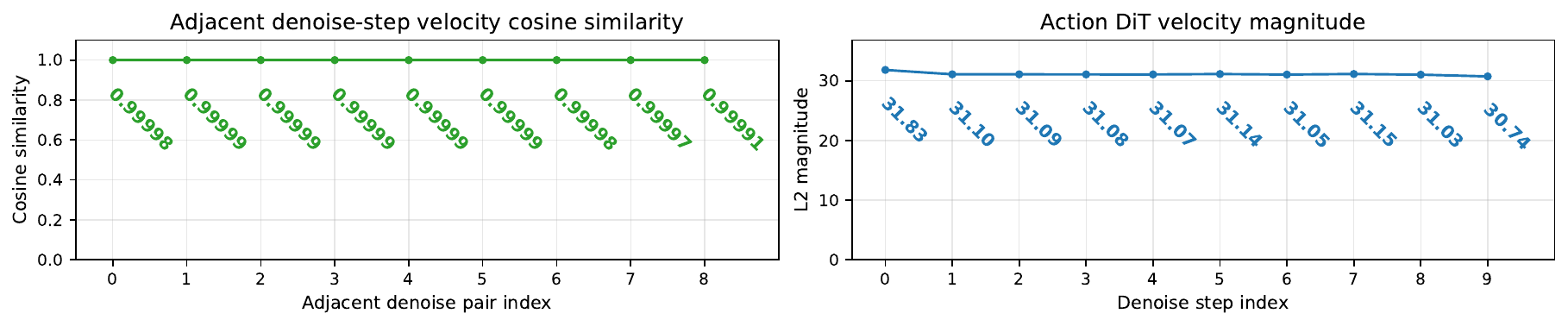}
    \caption{Left: cosine similarity between velocities predicted at adjacent denoising steps. Right: the $\ell_2$ magnitude of the predicted velocity at each step.}
    \label{fig:fastwam_velocity_analysis}
\end{figure*}

We examine whether the denoising trace of FastWAM, ImageWAM, and Kairos are consistent with the analysis in Section~\ref{subsec:Rethinking}.
Figure~\ref{fig:fastwam_velocity_analysis} reports the cosine similarity between predicted velocities  at adjacent denoising steps and their $\ell_2$ magnitudes for a representative action chunk. 
The adjacent step denoise velocity is extremely similar across denoising step, demonstrating the denoise trace is straight.
Corresponding results for ImageWAM and Kairos are provided in the appendix.

\subsection{Ablation Studies}
\label{sec:ablations}

\paragraph{Ablation of Bandwidth-Constrained Action Decoding.}
We evaluate the contribution of bandwidth-constrained action decoding by comparing the same calibrated checkpoint with and without it.
Table~\ref{tab:bandwidth_ablation} reports success before and after applying it, it improves 1Step-WAM success rates under both clean and randomized settings across all three backbones, with gains ranging from 0.7 to 2.0 percentage points, demonstrating the effectiveness of bandwidth-constrained action decoding.
We also apply the same projection to DMD2~\citep{yin2024improved}, SnapFlow~\citep{luan2026snapflow} and Flash-WAM~\citep{akbari2026flash} on FastWAM, both methods are improved, suggesting that the benefits extend beyond our approach to other distillation methods.

\begin{table}[!t]
    \centering
    \caption{
        Ablation of bandwidth-constrained action decoding on RoboTwin retaining first $K$ modes
    }
    \label{tab:bandwidth_ablation}
    \small
    \setlength{\tabcolsep}{5pt}
    \renewcommand{\arraystretch}{1.2}
    
    \begin{tabular}{@{}lllcc@{}}
        \toprule
        \textbf{Model} & \textbf{Method} & $K$ & \textbf{Clean} & \textbf{Randomized} \\
        \midrule
        \textbf{FastWAM}
        & Ours & 8
        & $89.64 \rightarrow \mathbf{91.10}$
        & $88.74 \rightarrow \mathbf{90.68}$ \\
        & DMD2~\citep{yin2024improved} & 8
        & $82.48 \rightarrow \mathbf{84.08}$
        & $83.40 \rightarrow \mathbf{83.60}$ \\
        & SnapFlow~\citep{luan2026snapflow} & 8
        & $90.52 \rightarrow \mathbf{91.32}$
        & $86.96 \rightarrow \mathbf{91.36}$ \\
        & Flash-WAM~\citep{akbari2026flash} & 8
        & $90.44 \rightarrow \mathbf{91.08}$
        & $89.80 \rightarrow \mathbf{91.00}$ \\
        \midrule
        \textbf{ImageWAM}
        & Ours & 4
        & $91.56 \rightarrow \mathbf{92.26}$
        & $91.72 \rightarrow \mathbf{92.62}$ \\
        \midrule
        \textbf{Kairos}
        & Ours & 8
        & $94.76 \rightarrow \mathbf{96.56}$
        & $92.24 \rightarrow \mathbf{93.84}$ \\
        \bottomrule
    \end{tabular}
\end{table}

\paragraph{Analysis of joint execution.}
Table~\ref{tab:joint_execution_ablation} isolates the effect of jointly scheduling video prefill and one-step action
generation. Compared with sequential execution, the proposed schedule reduces end-to-end latency by 39.65\%, 33.95\%, and
45.30\% on FastWAM, ImageWAM, and Kairos, respectively, corresponding to $1.66\times$, $1.51\times$, and $1.83\times$
speedups. 
\begin{table}[!t]
    \centering
    \caption{
        Ablation of joint video--action execution.
        ``w/o joint execution'' denotes  naive sequential execution , whereas
        ``w/ joint execution'' uses the proposed joint execution
        schedule.
    }
    \label{tab:joint_execution_ablation}
    \small
    \setlength{\tabcolsep}{6pt}
    \renewcommand{\arraystretch}{1.2}
    \begin{tabular}{@{}lrrrr@{}}
        \toprule
        \textbf{Model}
        & \textbf{w/o Joint (ms)} $\downarrow$
        & \textbf{w/ Joint (ms)} $\downarrow$
        & \textbf{Speedup} $\uparrow$
        & \textbf{Reduction} (\%) $\uparrow$ \\
        \midrule
        \textbf{FastWAM~\citep{yuan2026fastwam}}
        & 149.05 & 89.95 & $1.66\times$ & 39.65 \\
        \textbf{ImageWAM~\citep{zhang2026imagewam}}
        & 77.64 & 51.28 & $1.51\times$ & 33.95 \\
        \textbf{Kairos~\citep{kairosteam2026kairosregretawarenativeworldaction}}
        & 134.78 & 73.73 & $1.83\times$ & 45.30 \\
        \bottomrule
    \end{tabular}
\end{table}

  \section{Conclusion}
We propose 1Step-WAM, a lightweight framework for converting pretrained multi-step World Action Models into one-step action generators.
Motivated by the concentration of conditional action distributions and the approximately straight denoising trajectories, we introduced a small calibration head that corrects the initial velocity prediction while keeping the backbone frozen.
We further improved action quality through bandwidth-constrained decoding and propose a joint execution schedule to overlap visual prefill with action generation.
Evaluations on RoboTwin, LIBERO, and real-world manipulation tasks show that 1Step-WAM largely preserves multi-step task success with a single action-generation pass and with considerable speedup. 

  \newpage
  \section*{AI USE STATEMENT}
In this work, we utilized generative AI tools to assist with drafting, editing, and polishing text, as well as generating routine code for data analysis and plotting. The authors have meticulously reviewed and revised all AI-assisted text, and thoroughly tested all AI-generated code. We confirm that every reported result and metric originates entirely from our own physical and simulated experiments. The authors assume full responsibility for the final content, claims, and artifacts presented in this paper.

  \bibliography{iclr2027_conference}
  \bibliographystyle{iclr2027_conference}

  \section{Appendix}
\subsection{Detailed Derivation}
\label{app:velocity-derivation}

We provide the detailed velocity derivation underlying
Section~\ref{subsec:Rethinking}.

\subsubsection{Theoretical setup}
\label{app:velocity-assumptions}

\paragraph{Distributions and independent coupling.}
Let $d$ denote the action dimension. For fixed conditioning inputs,
we assume a Gaussian clean target action distribution and a Gaussian
noise prior:
\begin{equation}
    \bm x_0\sim\mathcal{N}(\bm \mu,\sigma^2 \bm I_d),
    \qquad \bm x_1\sim\mathcal{N}(\bm 0,\bm I_d),
    \qquad \bm x_0\perp \bm x_1,
\end{equation}
where $\bm \mu\in\mathbb{R}^d$ is the target action center and
$0<\sigma^2\ll1$ is the residual variance. The independence assumption
applies to the endpoints sampled for the training interpolation.

\paragraph{Linear interpolation and marginal distribution.}
The conditional flow-matching interpolation defines the forward
noising path:
\begin{equation}
    \bm x_t=(1-t)\bm x_0+t\bm x_1,\qquad t\in[0,1].
\end{equation}
Denoising traverses time in the reverse direction, from $t=1$ to $t=0$.
By independence and the linearity of Gaussian random variables,
\begin{equation}
\begin{aligned}
    \mathbb{E}[\bm x_t]
    &=(1-t)\mathbb{E}[\bm x_0]+t\mathbb{E}[\bm x_1]=(1-t)\bm \mu,\\
    \operatorname{Var}(\bm x_t)
    &=(1-t)^2\operatorname{Var}(\bm x_0)+t^2\operatorname{Var}(\bm x_1)\\
    &=\bigl[(1-t)^2\sigma^2+t^2\bigr]\bm I_d.
\end{aligned}
\end{equation}
Define the time-dependent variance and its derivative as
\begin{equation}
\begin{aligned}
    D(t)&=(1-t)^2\sigma^2+t^2,\\
    D'(t)&=-2(1-t)\sigma^2+2t
          =2\bigl[t-(1-t)\sigma^2\bigr].
\end{aligned}
\label{eq:app-variance}
\end{equation}
The marginal distribution is therefore
\begin{equation}
    \bm x_t\sim\mathcal{N}\bigl((1-t)\bm \mu,D(t)\bm I_d\bigr).
\end{equation}

\subsubsection{Derivation of the optimal velocity field}
\label{app:optimal-velocity-field}

The optimal flow-matching velocity field is defined with respect to
increasing $t$ as the conditional expectation of the interpolation
velocity:
\begin{equation}
    \bm v^*(\bm x_t,t)=\mathbb{E}[\bm x_1-\bm x_0\mid \bm x_t].
\end{equation}
For jointly Gaussian random vectors $\bm A$ and $\bm B$, the conditional
expectation identity is
\begin{equation}
    \mathbb{E}[\bm A\mid \bm B]
    =\bm \mu_A+\bm \Sigma_{AB}\bm \Sigma_{BB}^{-1}(\bm B-\bm \mu_B).
\end{equation}
We apply this identity with $\bm A=\bm x_1-\bm x_0$ and $\bm B=\bm x_t$.

\paragraph{Means and covariances.}
The means and the covariance of $\bm B$ are
\begin{equation}
\begin{aligned}
    \bm \mu_A&=\mathbb{E}[\bm x_1-\bm x_0]=-\bm \mu,\\
    \bm \mu_B&=\mathbb{E}[\bm x_t]=(1-t)\bm \mu,\\
    \bm \Sigma_{BB}&=\operatorname{Var}(\bm x_t)=D(t)\bm I_d.
\end{aligned}
\end{equation}
Using the bilinearity of covariance and $\bm x_0\perp \bm x_1$, the
cross-covariance is
\begin{equation}
\begin{aligned}
    \bm \Sigma_{AB}
    &=\operatorname{Cov}\bigl(\bm x_1-\bm x_0,(1-t)\bm x_0+t\bm x_1\bigr)\\
    &=\operatorname{Cov}(\bm x_1,t\bm x_1)
      -\operatorname{Cov}(\bm x_0,(1-t)\bm x_0)\\
    &=t\operatorname{Var}(\bm x_1)-(1-t)\operatorname{Var}(\bm x_0)\\
    &=\bigl[t-(1-t)\sigma^2\bigr]\bm I_d.
\end{aligned}
\end{equation}

\paragraph{Substitution and simplification.}
Substituting these quantities into the conditional expectation gives
\begin{equation}
\begin{aligned}
    \bm v^*(\bm x_t,t)
    &=-\bm \mu+\bigl[t-(1-t)\sigma^2\bigr]\bm I_d
       \bigl[D(t)\bm I_d\bigr]^{-1}\bigl[\bm x_t-(1-t)\bm \mu\bigr]\\
    &=-\bm \mu+\frac{t-(1-t)\sigma^2}{D(t)}\bigl[\bm x_t-(1-t)\bm \mu\bigr]\\
    &=\frac{\bigl\{-D(t)-(1-t)[t-(1-t)\sigma^2]\bigr\}\bm \mu
             +[t-(1-t)\sigma^2]\bm x_t}{D(t)}.
\end{aligned}
\end{equation}
The numerator of the coefficient of $\bm \mu$ simplifies to
\begin{equation}
\begin{aligned}
    -D(t)-(1-t)\bigl[t-(1-t)\sigma^2\bigr]
    &=-(1-t)^2\sigma^2-t^2-t(1-t)+(1-t)^2\sigma^2\\
    &=-t^2-t(1-t)\\
    &=-t\bigl[t+(1-t)\bigr]=-t.
\end{aligned}
\label{eq:app-variance-identity}
\end{equation}
We thus obtain the closed-form velocity field:
\begin{equation}
    \bm v^*(\bm x_t,t)
    =-\frac{t}{D(t)}\bm \mu
     +\frac{t-(1-t)\sigma^2}{D(t)}\bm x_t.
    \label{eq:app-optimal-field}
\end{equation}

\subsubsection{Analytical solution of the inference ODE}
\label{app:exact-trajectory}

During inference, we integrate the optimal velocity field backward
from $t=1$ to $t=0$, starting from a noise sample $\bm x_1$.
In this subsection, $\bm x_t$ denotes the ODE trajectory, which shares
the marginals of the training interpolation but couples its endpoints
through the ODE:
\begin{equation}
    \dot{\bm x}_t
    =\frac{t-(1-t)\sigma^2}{D(t)}\bm x_t
     -\frac{t}{D(t)}\bm \mu,
    \qquad \bm x(1)=\bm x_1.
    \label{eq:app-inference-ode}
\end{equation}
Here, $\dot{\bm x}_t$ is the derivative with respect to increasing $t$;
denoising updates use $\Delta t<0$.
Writing the equation as $\dot{\bm x}_t+P(t)\bm x_t=\bm Q(t)$ gives
\begin{equation}
    P(t)=-\frac{t-(1-t)\sigma^2}{D(t)}
        =-\frac{D'(t)}{2D(t)},
    \qquad \bm Q(t)=-\frac{t}{D(t)}\bm \mu.
\end{equation}

\paragraph{Integrating factor.}
The integrating factor is
\begin{equation}
\begin{aligned}
    I(t)
    &=\exp\left(\int_1^t P(\tau)\,d\tau\right)\\
    &=\exp\left(-\frac{1}{2}\int_1^t
           \frac{D'(\tau)}{D(\tau)}\,d\tau\right)\\
    &=\exp\left(-\frac{1}{2}\ln D(t)
               +\frac{1}{2}\ln D(1)\right).
\end{aligned}
\end{equation}
Since $D(1)=1$, this reduces to $I(t)=D(t)^{-1/2}$.
Multiplying the ODE by this factor yields
\begin{equation}
    D(t)^{-1/2}\dot{\bm x}_t
    -\frac{1}{2}D(t)^{-3/2}D'(t)\bm x_t
    =-\frac{t}{D(t)^{3/2}}\bm \mu.
\end{equation}
By the product rule, the left-hand side is the derivative of
$D(t)^{-1/2}\bm x_t$. Therefore,
\begin{equation}
    \frac{d}{dt}\left[\frac{\bm x_t}{\sqrt{D(t)}}\right]
    =-\frac{t}{D(t)^{3/2}}\bm \mu.
    \label{eq:app-integrating-factor}
\end{equation}

\paragraph{Antiderivative of the right-hand side.}
Differentiating $(1-t)/\sqrt{D(t)}$ gives
\begin{equation}
\begin{aligned}
    \frac{d}{dt}\left[\frac{1-t}{\sqrt{D(t)}}\right]
    &=\frac{-\sqrt{D(t)}
       -\frac{1}{2}(1-t)D(t)^{-1/2}D'(t)}{D(t)}\\
    &=\frac{-D(t)-\frac{1}{2}(1-t)D'(t)}{D(t)^{3/2}}.
\end{aligned}
\end{equation}
Substituting $D(t)$ and $D'(t)$ into the numerator,
\begin{equation}
\begin{aligned}
    -D(t)-\frac{1}{2}(1-t)D'(t)
    &=-(1-t)^2\sigma^2-t^2-(1-t)\bigl[t-(1-t)\sigma^2\bigr]\\
    &=-t^2-t(1-t)=-t.
\end{aligned}
\end{equation}
Hence
\begin{equation}
    \frac{d}{dt}\left[\frac{1-t}{\sqrt{D(t)}}\right]
    =-\frac{t}{D(t)^{3/2}}.
\end{equation}

\paragraph{Inference trajectory and instantaneous velocity.}
Integrating Equation~\eqref{eq:app-integrating-factor} from $1$ to $t$
gives
\begin{equation}
    \int_1^t\frac{d}{d\tau}
       \left[\frac{\bm x_\tau}{\sqrt{D(\tau)}}\right]d\tau
    =-\int_1^t\frac{\tau}{D(\tau)^{3/2}}\,d\tau\cdot\bm \mu,
\end{equation}
and therefore
\begin{equation}
    \frac{\bm x_t}{\sqrt{D(t)}}-\frac{\bm x_1}{\sqrt{D(1)}}
    =\left[\frac{1-\tau}{\sqrt{D(\tau)}}\right]_1^t\bm \mu.
\end{equation}
Using $D(1)=1$ and multiplying by $\sqrt{D(t)}$, we obtain
\begin{equation}
    \bm x_t=\sqrt{D(t)}\,\bm x_1+(1-t)\bm \mu
       =\sqrt{(1-t)^2\sigma^2+t^2}\,\bm x_1+(1-t)\bm \mu.
    \label{eq:app-exact-trajectory}
\end{equation}
At $t=0$, the endpoint is $\bm x_{t=0}=\bm \mu+\sigma \bm x_1$, which follows
the target distribution $\mathcal{N}(\bm \mu,\sigma^2 \bm I_d)$.
Differentiation with respect to $t$ gives the instantaneous velocity:
\begin{equation}
    \dot{\bm x}_t=-\bm \mu+C(t)\bm x_1,
    \qquad
    C(t)=\frac{D'(t)}{2\sqrt{D(t)}}
        =\frac{t-(1-t)\sigma^2}
               {\sqrt{(1-t)^2\sigma^2+t^2}}.
    \label{eq:app-exact-velocity}
\end{equation}

\subsubsection{Main-interval collinearity and terminal boundary behavior}
\label{app:velocity-geometry}

\paragraph{Theorem 1.}
For any fixed $\epsilon\in(0,1)$ independent of $\sigma$, as
$\sigma\to0^+$, $C(t)$ converges uniformly to $1$ on $[\epsilon,1]$.
For a fixed noise realization $\bm x_1$, the ODE velocity therefore
approaches $\bm x_1-\bm \mu$ throughout this interval, and the reverse-time
denoising velocity $-\dot{\bm x}_t$ approaches $\bm \mu-\bm x_1$.
Near the clean endpoint, the coefficient crosses zero at
$t^*=\sigma^2/(1+\sigma^2)$, where $\dot{\bm x}_{t^*}=-\bm \mu$.

\paragraph{Main-interval collinearity.}
For $t\in[\epsilon,1]$, we have $t\geq\epsilon>0$ and can write
\begin{equation}
    C(t)
    =\frac{1-\frac{(1-t)\sigma^2}{t}}
           {\sqrt{1+\left(\frac{(1-t)\sigma}{t}\right)^2}}.
\end{equation}
The two ratios satisfy the uniform bounds
\begin{equation}
    0\leq\frac{(1-t)\sigma}{t}
       \leq\frac{(1-\epsilon)\sigma}{\epsilon}
       =\mathcal{O}(\sigma),
    \qquad
    0\leq\frac{(1-t)\sigma^2}{t}
       \leq\frac{(1-\epsilon)\sigma^2}{\epsilon}
       =\mathcal{O}(\sigma^2).
\end{equation}
The constants in these bounds may depend on $\epsilon$ but not on $t$
or $\sigma$. Consequently,
\begin{equation}
    C(t)=\frac{1+\mathcal{O}(\sigma^2)}
                  {\sqrt{1+\mathcal{O}(\sigma^2)}},
    \qquad
    \lim_{\sigma\to0^+}
    \sup_{t\in[\epsilon,1]}|C(t)-1|=0.
\end{equation}
It follows that $\dot{\bm x}_t=-\bm \mu+C(t)\bm x_1$ converges uniformly to
$\bm x_1-\bm \mu$ on this interval. For $\bm x_1-\bm \mu\neq\bm 0$ and any
$t_1,t_2\in[\epsilon,1]$, the velocity cosine similarity satisfies
\begin{equation}
    \lim_{\sigma\to0^+}\cos(\dot{\bm x}_{t_1},\dot{\bm x}_{t_2})
    =\frac{(\bm x_1-\bm \mu)^\top(\bm x_1-\bm \mu)}
           {\|\bm x_1-\bm \mu\|_2\,\|\bm x_1-\bm \mu\|_2}=1.
\end{equation}
Negating both velocities leaves this cosine similarity unchanged,
so the same collinearity holds in the denoising direction.

\paragraph{Terminal transition scale.}
The main-interval convergence does not extend uniformly to $t=0$,
since $C(0)=-\sigma\to0$ rather than $1$. To resolve the terminal
transition, set $t=\alpha\sigma$ for fixed $\alpha\geq0$. Then
\begin{equation}
    C(\alpha\sigma)
    =\frac{\alpha-(1-\alpha\sigma)\sigma}
           {\sqrt{\alpha^2+(1-\alpha\sigma)^2}}
    \longrightarrow\frac{\alpha}{\sqrt{1+\alpha^2}}.
\end{equation}
This limiting profile describes the change in the noise coefficient
on the scale $t=\mathcal{O}(\sigma)$ near the clean endpoint,
consistent with Section~\ref{subsec:Rethinking}.

\paragraph{Terminal zero crossing.}
Solving $C(t^*)=0$ gives
\begin{equation}
    t^*-(1-t^*)\sigma^2=0
    \quad\Longrightarrow\quad
    t^*=\frac{\sigma^2}{1+\sigma^2}
       =\sigma^2+\mathcal{O}(\sigma^4).
    \label{eq:app-zero-crossing}
\end{equation}
Thus the zero crossing lies $\mathcal{O}(\sigma^2)$ from the clean
endpoint $t=0$.
At this time the noise term in the instantaneous velocity vanishes:
\begin{equation}
    C(t^*)=0,\qquad \dot{\bm x}_{t^*}=-\bm \mu.
\end{equation}
For $t\in(t^*,1]$, $C(t)>0$, whereas for $t\in[0,t^*)$, $C(t)<0$.
At the clean endpoint,
\begin{equation}
    C(0)=-\frac{\sigma^2}{\sqrt{\sigma^2}}=-\sigma,
    \qquad \dot{\bm x}_{t=0}=-\bm \mu-\sigma \bm x_1.
\end{equation}
The reverse-time denoising velocity therefore approaches $\bm \mu$ at
the clean endpoint as $\sigma\to0^+$. The scale
$\mathcal{O}(\sigma^2)$ specifies the zero-crossing location within
the broader $\mathcal{O}(\sigma)$ terminal transition region.

\paragraph{One-step estimate and residual error.}
At the noise endpoint, $D(1)=1$ and $\bm v^*(\bm x_1,1)=\bm x_1-\bm \mu$.
A single reverse-time Euler step with $\Delta t=-1$ therefore gives
\begin{equation}
    \widehat{\bm x}_0=\bm x_1-\bm v^*(\bm x_1,1)=\bm \mu.
\end{equation}
For the same initial noise, the exact ODE endpoint is
$\bm x_{t=0}=\bm \mu+\sigma \bm x_1$. Hence, under this Gaussian model and
the optimal velocity field,
\begin{equation}
    \widehat{\bm x}_0-\bm x_{t=0}=-\sigma \bm x_1,
    \qquad
    \mathbb{E}_{\bm x_1}\bigl\|\widehat{\bm x}_0-\bm x_{t=0}\bigr\|_2^2=d\sigma^2.
\end{equation}
Thus the one-step estimate recovers the conditional mean and
approximates the ODE endpoint as $\sigma\to0^+$ for fixed $d$;
it does not reproduce the residual output variance at finite $\sigma$.

\subsection{Per-Joint Spectral Analysis}
  \label{app:action_spectra}

\begin{figure}[htbp]
      \centering
      \includegraphics[width=\textwidth]
          {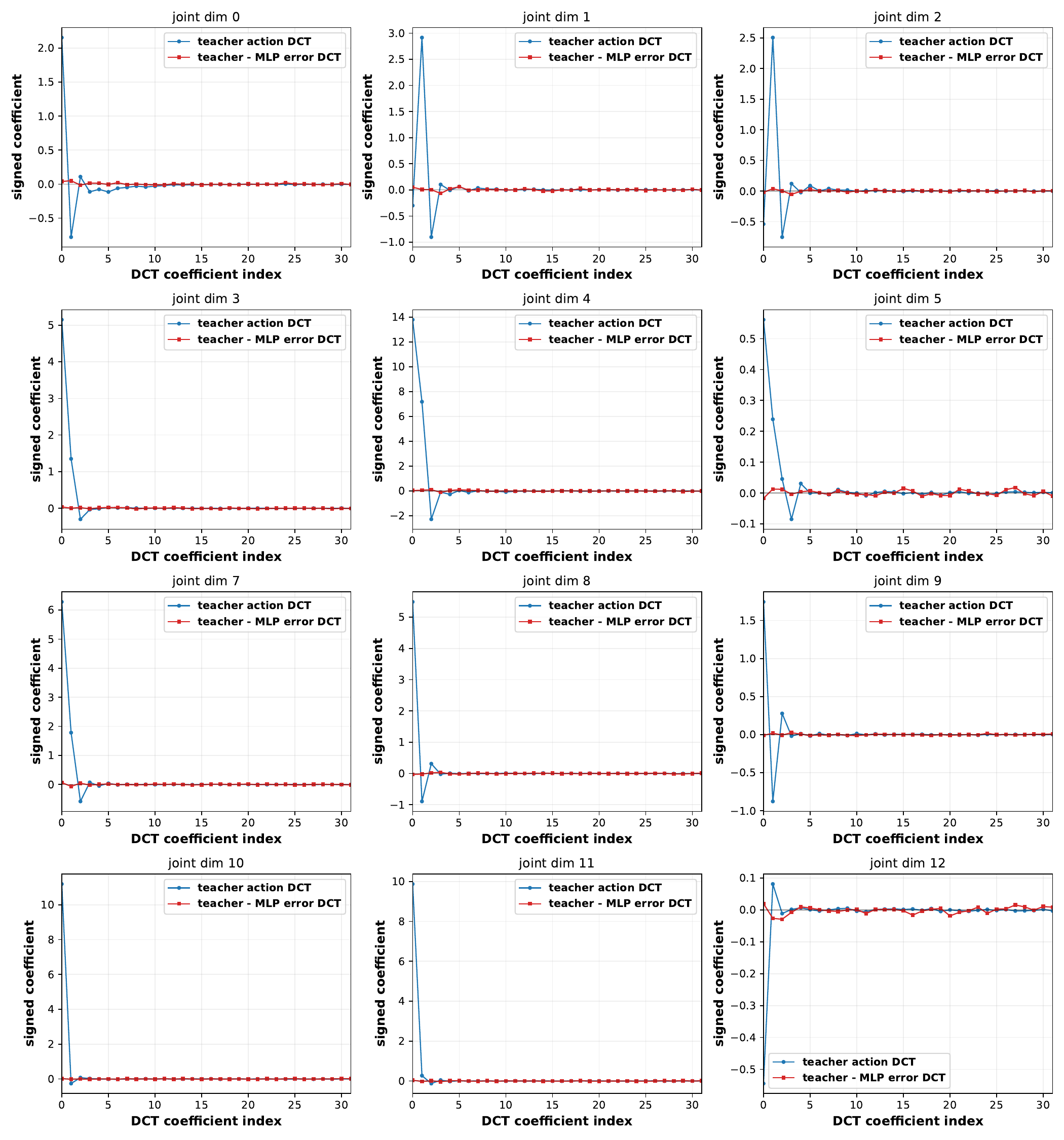}
      \caption{
          Per-joint temporal DCT coefficients of FastWAM
          teacher actions (blue) and teacher-minus-prediction
          residuals from the calibrated one-step model (red).
          Each panel corresponds to one continuous joint
          dimension, with frequency increasing along the
          horizontal axis.
      }
      \label{fig:fastwam_dct_compare}
  \end{figure}

  Figure~\ref{fig:fastwam_dct_compare} compares the temporal
  DCT coefficients of FastWAM teacher actions and the residual
  errors of the calibrated one-step predictions.
  The transform is applied independently along the action
  horizon for each continuous joint dimension.
  The blue curves show the teacher coefficients, while the
  red curves show the coefficients of the teacher-minus-prediction
  residual before bandwidth-constrained decoding.
  Since the plotted coefficients are signed, we compare their
  absolute magnitudes; the energy of each mode is given by
  the squared coefficient.

The teacher coefficients are largest in the low-frequency modes and become small
  at higher frequencies.
  Residual errors nevertheless remain visible in some of
  these higher-frequency modes.
  In particular, joint dimensions 5 and 12 exhibit modes
  where the error magnitude exceeds that of the corresponding teacher component.
  Prediction errors can therefore overwhelm weak high-frequency teacher signals, even when these errors are small relative to the dominant low-frequency components.

  This observation motivates retaining the first $K$ DCT
  modes to preserve the dominant teacher components while
  suppressing residual errors in the discarded frequencies.
  As established in our projection analysis, truncation
  reduces teacher-matching error when the total discarded
  error energy exceeds the total discarded teacher energy.

\subsection{Real Tasks Rollout on Pick Place and Hang Mug}
\begin{figure*}[htbp]
    \centering
    \includegraphics[width=\textwidth]{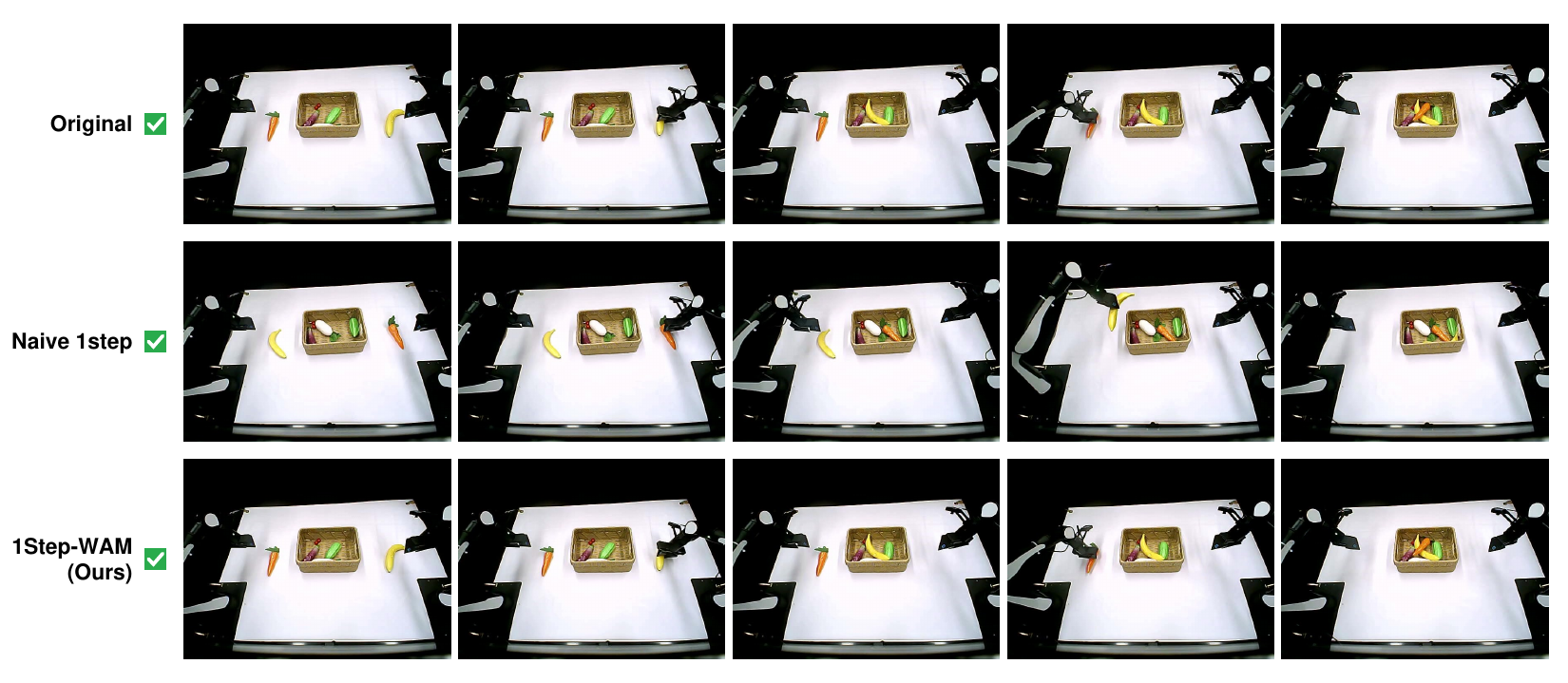}
    \caption{Real-world execution comparison on pick place. In the illustrated rollouts, both are success.}
    \label{fig:real_world_tasks_pick_place}
\end{figure*}

\begin{figure*}[htbp]
    \centering
    \includegraphics[width=\textwidth]{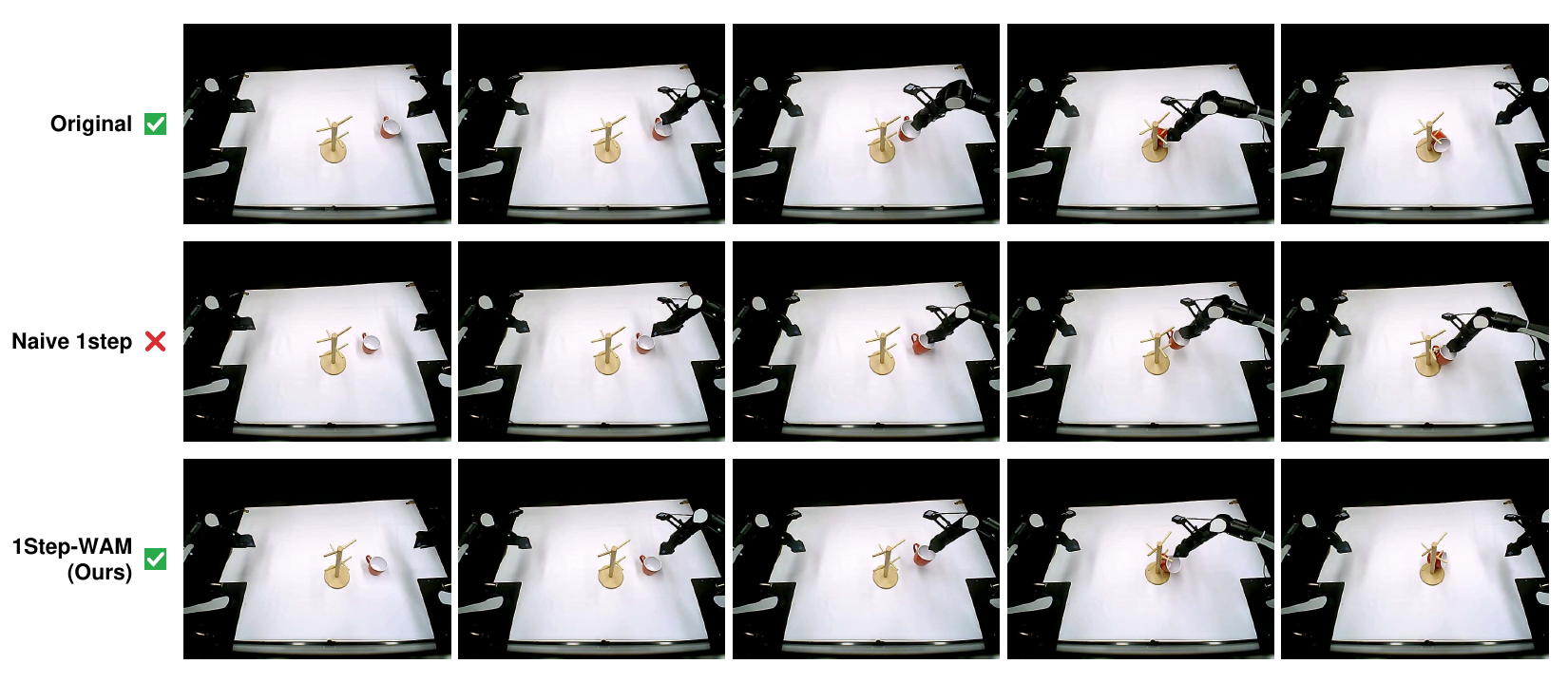}
    \caption{Real-world execution comparison on hang mug. In the illustrated rollouts, original FastWAM and 1Step-WAM successfully hang mug, while naive 1-step fails.}
    \label{fig:real_world_tasks_hang_mug}
\end{figure*}

The real execution sequences of the three methods on Pick Place and Hang Mug are shown as Figure~\ref{fig:real_world_tasks_pick_place} and Figure~\ref{fig:real_world_tasks_hang_mug}.

\subsection{Empirical Analysis of Denoising Trajectories on ImageWAM and Kairos}
\begin{figure*}[htbp]
    \centering
    \includegraphics[width=\textwidth]{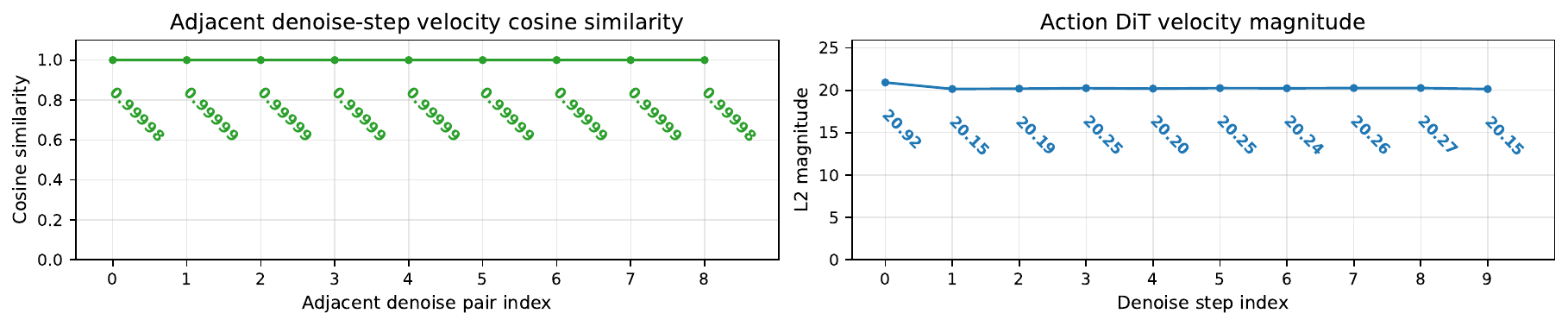}
    \caption{Left: cosine similarity between velocities predicted at adjacent denoising steps. Right: the $\ell_2$ magnitude of the predicted velocity at each step.}
    \label{fig:imagewam_velocity_analysis}
\end{figure*}

\begin{figure*}[htbp]
    \centering
    \includegraphics[width=\textwidth]{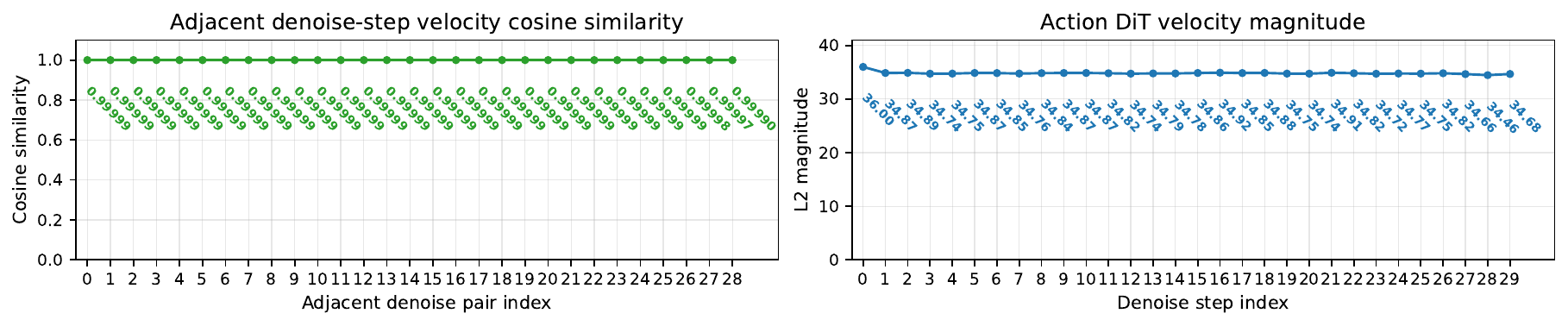}
    \caption{Left: cosine similarity between velocities predicted at adjacent denoising steps. Right: the $\ell_2$ magnitude of the predicted velocity at each step.}
    \label{fig:kairos_velocity_analysis}
\end{figure*}

Figure~\ref{fig:imagewam_velocity_analysis} and Figure~\ref{fig:kairos_velocity_analysis} reports the cosine similarity between predicted velocities at adjacent denoising steps and their $\ell_2$ magnitudes for a representative action chunk. Both adjacent step denoise velocity is extremely similar across denoising step, demonstrating the denoise trace is straight.


\subsection{Experimental Details}
In this section more details of the experiments are provided.

\subsubsection{Calibration Head Configuration}
We introduce a velocity-residual calibration head to refine the one-step action DiT prediction. The calibration for the \(i\text{-th}\) action token is explicitly conditioned on three core components: the final action DiT block output hidden state \(\bm h_1^{(i)} \in \mathbb{R}^{1024}\), the first step predicted velocity \(\bm v_1^{(i)} \in \mathbb{R}^{d_a}\), and the noised action token \(\bm x_1^{(i)}\), which is  augmented \(\bm x_1^{(i)}\) with the feature \(\bm e^{(i)} \in \mathbb{R}^{1024}\) produced by the frozen action DiT encoder($\bm e=\textit{ActionDit.Encoder}(\bm x_1)$) and a learnable horizon embedding \(\bm p^{(i)} \in \mathbb{R}^{128}\). Here \(d_a=7\) in Libero~\citep{liu2023libero} and \(d_a=14\) in RoboTwin~\citep{chen2025robotwin}.

These three core components are first projected into a compact feature space. Specifically, the action token is projected and seamlessly concatenated with its context embeddings to form a unified token representation \(\tilde{\bm x}_1^{(i)}\):
\[
\begin{aligned}
\tilde{\bm h}_1^{(i)} &= \bm W_h \mathrm{LN}(\bm h_1^{(i)}) + \bm b_h, && \bm W_h \in \mathbb{R}^{1024 \times 1024}, \\
\tilde{\bm v}_1^{(i)} &= \bm W_v \bm v_1^{(i)} + \bm b_v, && \bm W_v \in \mathbb{R}^{256 \times d_a}, \\
\tilde{\bm x}_1^{(i)} &= [\bm W_x \bm x_1^{(i)} + \bm b_x; \; \bm W_e \bm e^{(i)} + \bm b_e; \; \bm p^{(i)}], && \tilde{\bm x}_1^{(i)} \in \mathbb{R}^{640},
\end{aligned}
\]
where \(\bm W_x \in \mathbb{R}^{256 \times d_a}\) and \(\bm W_e \in \mathbb{R}^{256 \times 1024}\). 

The projected features are then concatenated together:
\[
\bm z^{(i)} = [\tilde{\bm h}_1^{(i)}; \tilde{\bm x}_1^{(i)}; \tilde{\bm v}_1^{(i)}] \in \mathbb{R}^{1920}.
\]

The velocity correction is finally predicted by a three-layer MLP:
\[
\begin{aligned}
\bm u^{(i)} &= \mathrm{GELU}(\bm W_1 \bm z^{(i)} + \bm b_1), \\
\bm q^{(i)} &= \mathrm{GELU}(\bm W_2 \bm u^{(i)} + \bm b_2), \\
\Delta \bm v^{(i)} &= \bm W_3 \bm q^{(i)} + \bm b_3.
\end{aligned}
\]
where
\[
\begin{aligned}
\bm W_1 \in \mathbb{R}^{4096 \times 1920}, \quad
\bm W_2 \in \mathbb{R}^{1024 \times 4096}, \quad
\bm W_3 \in \mathbb{R}^{d_a \times 1024}.
\end{aligned}
\]
Thus, the MLP has the width configuration
\[
1920 \rightarrow 4096 \rightarrow 1024 \rightarrow d_a.
\]

The calibrated action endpoint is then obtained by correcting the raw velocity:
\[
\hat{\bm a}_0 = \bm x_1 - \left(\bm v_1 + \Delta \bm v\right).
\]
The weights and bias of the final projection \((\bm W_3, \bm b_3)\) are initialized to zero. Consequently, before calibration learning, \(\Delta \bm v = \bm 0\), and the calibration head exactly preserves the original one-step action DiT prediction.

\subsection{Detailed RoboTwin result}
Per-task detailed result of our proposed 1Step-WAM is shown as Table~\ref{tab:fastwam-comparison}

\begin{table*}[htbp]
\centering
\caption{Comparison of Fast-WAM, ImageWAM, and Kairos across manipulation tasks.}
\label{tab:fastwam-comparison}
\resizebox{\textwidth}{!}{%
\begin{tabular}{lcccccccccccc}
\toprule
\multicolumn{1}{c}{Task} & \multicolumn{4}{c}{Fast-WAM} & \multicolumn{4}{c}{ImageWAM} & \multicolumn{4}{c}{Kairos} \\
\cmidrule(lr){2-5} \cmidrule(lr){6-9} \cmidrule(lr){10-13}
 & \multicolumn{2}{c}{Clean} & \multicolumn{2}{c}{Rand.} & \multicolumn{2}{c}{Clean} & \multicolumn{2}{c}{Rand.} & \multicolumn{2}{c}{Clean} & \multicolumn{2}{c}{Rand.} \\
 & Original & Ours & Original & Ours & 10-steps & Ours & 10-steps & Ours & Original & Ours & Original & Ours \\
\midrule
adjust\_bottle & 1.00 & 1.0 & 1.00 & 0.99 & 1.00 & 1.0 & 0.99 & 0.99 & 1.0 & 1.0 & 1.0 & 1.0 \\
beat\_block\_hammer & 0.99 & 1.0 & 0.97 & 0.98 & 1.00 & 0.98 & 0.99 & 0.9 & 1.0 & 1.0 & 0.98 & 1.0 \\
blocks\_ranking\_rgb & 1.00 & 1.0 & 1.00 & 0.98 & 0.99 & 0.9 & 0.99 & 0.98 & 1.0 & 1.0 & 1.0 & 1.0 \\
blocks\_ranking\_size & 0.94 & 0.91 & 0.98 & 0.92 & 0.95 & 0.9 & 0.98 & 0.96 & 1.0 & 1.0 & 0.92 & 0.96 \\
click\_alarmclock & 1.00 & 1.0 & 1.00 & 1.0 & 1.00 & 0.99 & 1.00 & 1.0 & 1.0 & 1.0 & 1.0 & 1.0 \\
click\_bell & 1.00 & 1.0 & 1.00 & 1.0 & 1.00 & 1.0 & 1.00 & 1.0 & 1.0 & 1.0 & 1.0 & 1.0 \\
dump\_bin\_bigbin & 0.97 & 0.93 & 0.96 & 0.96 & 0.95 & 0.99 & 0.92 & 0.95 & 0.92 & 0.98 & 0.96 & 0.98 \\
grab\_roller & 1.00 & 1.0 & 1.00 & 1.0 & 1.00 & 1.0 & 1.00 & 1.0 & 1.0 & 1.0 & 1.0 & 1.0 \\
handover\_block & 0.95 & 0.86 & 0.81 & 0.78 & 0.95 & 0.97 & 0.95 & 0.93 & 1.0 & 0.98 & 0.96 & 0.92 \\
handover\_mic & 0.99 & 0.95 & 1.00 & 0.97 & 0.99 & 1.0 & 0.99 & 1.0 & 0.98 & 0.96 & 0.96 & 1.0 \\
hanging\_mug & 0.58 & 0.66 & 0.62 & 0.56 & 0.77 & 0.64 & 0.73 & 0.62 & 0.98 & 0.96 & 0.62 & 0.56 \\
lift\_pot & 1.00 & 0.99 & 1.00 & 1.0 & 1.00 & 0.99 & 1.00 & 1.0 & 1.0 & 1.0 & 1.0 & 1.0 \\
move\_can\_pot & 0.90 & 0.93 & 0.88 & 0.92 & 0.95 & 0.88 & 0.97 & 0.87 & 0.98 & 0.98 & 0.98 & 0.92 \\
move\_pillbottle\_pad & 1.00 & 0.98 & 0.99 & 0.98 & 0.99 & 0.98 & 1.00 & 1.0 & 1.0 & 1.0 & 1.0 & 0.98 \\
move\_playingcard\_away & 1.00 & 1.0 & 1.00 & 1.0 & 1.00 & 1.0 & 0.99 & 0.99 & 1.0 & 0.98 & 1.0 & 1.0 \\
move\_stapler\_pad & 0.77 & 0.72 & 0.64 & 0.67 & 0.63 & 0.62 & 0.70 & 0.67 & 0.94 & 0.9 & 0.94 & 0.9 \\
open\_laptop & 0.98 & 0.96 & 1.00 & 0.99 & 0.98 & 0.97 & 0.99 & 0.98 & 1.0 & 0.88 & 1.0 & 0.92 \\
open\_microwave & 0.62 & 0.74 & 0.45 & 0.58 & 0.96 & 0.98 & 0.96 & 0.99 & 0.88 & 0.96 & 0.98 & 1.0 \\
pick\_diverse\_bottles & 0.80 & 0.88 & 0.85 & 0.88 & 0.85 & 0.83 & 0.89 & 0.89 & 0.96 & 0.94 & 0.94 & 0.94 \\
pick\_dual\_bottles & 1.00 & 0.99 & 0.96 & 0.96 & 0.99 & 0.91 & 0.99 & 0.92 & 1.0 & 1.0 & 1.0 & 0.96 \\
place\_a2b\_left & 0.95 & 0.94 & 0.93 & 0.95 & 0.95 & 0.94 & 0.96 & 0.96 & 1.0 & 1.0 & 0.96 & 0.92 \\
place\_a2b\_right & 0.93 & 0.95 & 0.99 & 0.97 & 0.98 & 0.97 & 0.96 & 0.96 & 1.0 & 1.0 & 1.0 & 1.0 \\
place\_bread\_basket & 0.91 & 0.94 & 0.93 & 0.93 & 0.96 & 0.96 & 0.91 & 0.91 & 1.0 & 0.96 & 0.96 & 0.96 \\
place\_bread\_skillet & 0.90 & 0.88 & 0.93 & 0.91 & 0.87 & 0.91 & 0.89 & 0.88 & 0.92 & 0.92 & 0.94 & 0.82 \\
place\_burger\_fries & 0.96 & 0.97 & 0.99 & 0.97 & 0.98 & 0.97 & 1.00 & 1.0 & 0.96 & 0.98 & 0.98 & 0.98 \\
place\_can\_basket & 0.71 & 0.66 & 0.69 & 0.69 & 0.74 & 0.79 & 0.66 & 0.65 & 0.86 & 0.84 & 0.64 & 0.56 \\
place\_cans\_plasticbox & 0.99 & 0.97 & 0.96 & 0.97 & 0.99 & 1.0 & 0.96 & 0.99 & 0.98 & 1.0 & 0.98 & 0.98 \\
place\_container\_plate & 0.96 & 1.0 & 1.00 & 1.0 & 0.98 & 0.96 & 0.98 & 1.0 & 1.0 & 1.0 & 1.0 & 1.0 \\
place\_dual\_shoes & 0.94 & 0.84 & 0.88 & 0.87 & 0.81 & 0.82 & 0.85 & 0.88 & 0.9 & 0.96 & 0.92 & 0.9 \\
place\_empty\_cup & 1.00 & 0.99 & 1.00 & 1.0 & 1.00 & 1.0 & 1.00 & 1.0 & 1.0 & 1.0 & 1.0 & 1.0 \\
place\_fan & 0.96 & 0.97 & 0.96 & 0.89 & 0.97 & 0.94 & 0.94 & 0.94 & 1.0 & 1.0 & 0.96 & 0.92 \\
place\_mouse\_pad & 0.83 & 0.74 & 0.89 & 0.81 & 0.86 & 0.88 & 0.90 & 0.87 & 1.0 & 0.98 & 1.0 & 1.0 \\
place\_object\_basket & 0.89 & 0.81 & 0.88 & 0.81 & 0.74 & 0.85 & 0.77 & 0.79 & 0.86 & 0.88 & 0.88 & 0.94 \\
place\_object\_scale & 0.90 & 0.92 & 0.97 & 0.95 & 0.96 & 0.96 & 0.96 & 0.96 & 1.0 & 0.98 & 0.96 & 0.98 \\
place\_object\_stand & 0.90 & 0.94 & 0.94 & 0.95 & 0.98 & 0.96 & 0.95 & 0.97 & 0.98 & 0.98 & 0.96 & 0.92 \\
place\_phone\_stand & 0.97 & 0.98 & 0.99 & 0.98 & 1.00 & 0.99 & 1.00 & 0.97 & 1.0 & 1.0 & 1.0 & 0.96 \\
place\_shoe & 0.96 & 0.97 & 0.99 & 0.99 & 0.97 & 0.97 & 0.97 & 0.98 & 1.0 & 1.0 & 0.98 & 1.0 \\
press\_stapler & 0.90 & 0.91 & 0.97 & 0.94 & 0.97 & 0.95 & 1.00 & 1.0 & 1.0 & 1.0 & 0.94 & 0.94 \\
put\_bottles\_dustbin & 0.95 & 0.91 & 0.90 & 0.82 & 0.97 & 0.97 & 0.90 & 0.92 & 0.96 & 0.94 & 0.92 & 0.98 \\
put\_object\_cabinet & 0.94 & 0.82 & 0.89 & 0.85 & 0.91 & 0.92 & 0.91 & 0.92 & 0.84 & 0.88 & 0.9 & 0.9 \\
rotate\_qrcode & 0.93 & 0.86 & 0.89 & 0.89 & 0.88 & 0.88 & 0.92 & 0.9 & 0.9 & 0.84 & 0.96 & 0.94 \\
scan\_object & 0.89 & 0.84 & 0.92 & 0.84 & 0.95 & 0.89 & 0.89 & 0.94 & 0.94 & 1.0 & 0.94 & 0.92 \\
shake\_bottle & 1.00 & 1.0 & 1.00 & 1.0 & 1.00 & 1.0 & 1.00 & 1.0 & 1.0 & 1.0 & 1.0 & 1.0 \\
shake\_bottle\_horizontally & 1.00 & 1.0 & 1.00 & 1.0 & 1.00 & 1.0 & 1.00 & 1.0 & 1.0 & 1.0 & 1.0 & 1.0 \\
stack\_blocks\_three & 0.95 & 0.93 & 0.97 & 0.96 & 0.96 & 0.92 & 0.99 & 0.96 & 0.98 & 0.98 & 0.96 & 0.94 \\
stack\_blocks\_two & 1.00 & 1.0 & 1.00 & 1.0 & 1.00 & 0.98 & 0.99 & 0.98 & 1.0 & 1.0 & 1.0 & 1.0 \\
stack\_bowls\_three & 0.80 & 0.84 & 0.81 & 0.83 & 0.80 & 0.8 & 0.79 & 0.82 & 0.82 & 0.78 & 0.84 & 0.9 \\
stack\_bowls\_two & 0.92 & 0.96 & 0.98 & 0.95 & 0.97 & 0.95 & 0.96 & 0.97 & 0.98 & 1.0 & 0.98 & 1.0 \\
stamp\_seal & 0.90 & 0.89 & 0.94 & 0.88 & 0.83 & 0.73 & 0.83 & 0.81 & 1.0 & 0.98 & 0.88 & 0.82 \\
turn\_switch & 0.61 & 0.62 & 0.59 & 0.62 & 0.76 & 0.74 & 0.78 & 0.74 & 0.96 & 0.88 & 0.8 & 0.7 \\
\midrule
Mean & 0.9188 & 0.9110 & 0.9178 & 0.9068 & 0.9338 & 0.9226 & 0.9330 & 0.9262 & 0.9696 & 0.9656 & 0.9496 & 0.9384 \\
\bottomrule
\end{tabular}%
}
\end{table*}

  \end{document}